\documentclass{article}

\usepackage{arxiv}

\usepackage{newtxtext}
\usepackage{newtxmath}

\usepackage[utf8]{inputenc}
\usepackage[T2A,T1]{fontenc}
\usepackage[english]{babel}

\usepackage{hyperref}
\usepackage{url}
\usepackage{booktabs}
\usepackage{amsfonts}
\usepackage{nicefrac}
\usepackage{microtype}
\usepackage{cleveref}
\usepackage{graphicx}
\usepackage{natbib}
\usepackage{doi}
\usepackage{authblk}

\title{Large Language Models for Low-Resource Languages: \\
       A Conceptual Framework for an Electronic Explanatory Dictionary \\
       of the Tajik Language}

\author[1]{M.~K.~Arabov\thanks{Email: \texttt{MKArabov@kpfu.ru}}}
\author[2]{S. ~M. ~Pirzoda \thanks{Email: \texttt{3samr@list.ru}}}
\author[2]{B. ~A.~Sultonov\thanks{Email: \texttt{sbehruz40@gmail.com}}}

\affil[1]{Kazan Federal University, Institute of Computational Mathematics and Information Technologies, Kazan, Russia}
\affil[2]{Tajik National University, Department of Information and Communication Technologies, Faculty of Mechanics and Mathematics, Dushanbe, Tajikistan}

\renewcommand{\shorttitle}{LLMs for Tajik Dictionary}

\hypersetup{
    pdftitle={Large Language Models for Low-Resource Languages: A Conceptual Framework for an Electronic Explanatory Dictionary of the Tajik Language},
    pdfsubject={cs.CL, cs.AI},
    pdfauthor={M. K. Arabov, B. Sultonov, Pirzoda S. S. },
    pdfkeywords={large language models, low-resource languages, Tajik language, electronic dictionary, natural language processing, neural approaches},
}

\newcommand{\tj}[1]{{\fontencoding{T2A}\selectfont #1}}

\begin{document}

\maketitle


\begin{abstract}
This paper presents a conceptual framework for developing an electronic explanatory dictionary of the Tajik language using large language models (LLMs). The relevance of the work stems from the absence of a comprehensive digital lexicographic resource for Tajik that is comparable in functionality to dictionaries for high-resource languages, and from the limited adaptation of modern natural language processing technologies to low-resource language systems. Based on a systematic survey of existing linguistic, statistical, and corpus resources, we propose a dictionary architecture that integrates modules for morphological analysis, lemmatization, semantic clustering, and dictionary entry generation using LLMs. The choice of subword tokenization is justified by the agglutinative nature of Tajik morphology and its high morphological variability, along with a parameter-efficient fine-tuning (PEFT) strategy suitable for limited annotated data. The novelty of the work lies in proposing the first holistic conceptual architecture of an explanatory dictionary for Tajik that unifies classical lexicographic methods, language statistics, and generative capabilities of LLMs into a single system. The practical significance of the study is the formation of a methodological foundation for developing a full-featured electronic dictionary that can serve both as a lexicographic tool and as a core resource for machine translation, automatic summarization, sentiment analysis, and other applied NLP tasks. The paper is intended for specialists in computational linguistics, lexicography, and developers of natural language processing systems working with low-resource languages.
\end{abstract}

\noindent\textit{Keywords:} large language models, electronic explanatory dictionary, Tajik language, low-resource languages, parameter-efficient fine-tuning, subword tokenization, lexicography, morphological analysis, semantic clustering, corpus linguistics, natural language processing.


\section{Introduction}

The rapid development of natural language processing (NLP) technologies in recent years has led to a qualitative shift in the creation of digital linguistic resources. Particularly notable results have been achieved based on large language models (LLMs), which have demonstrated high effectiveness in tasks of text generation, understanding, and analysis for languages with a high level of digital support. However, this progress is extremely uneven: languages that lack sufficient annotated corpora, computational infrastructure, and research communities remain virtually excluded from modern LLM ecosystems \citep{arabov2025developing}. The Tajik language is among such low-resource languages.

Tajik, being the state language of the Republic of Tajikistan and possessing a centuries-old literary tradition, still lacks a comprehensive electronic explanatory dictionary comparable in functionality to lexicographic resources for world languages. Existing explanatory dictionaries of Tajik, available in printed form \citep{shukurov2008tajik} or as limited electronic versions, do not meet modern requirements for digital lexicographic resources: they do not support dynamic updating, are not integrated with automatic text processing systems, and do not employ machine learning methods for generating and updating dictionary entries.

At the same time, over the past two decades, a scientific school in computational linguistics has been established in Tajikistan, led by Z.D. Usmanov and his students, within which fundamental foundations for automatic text processing in Tajik have been developed. Conceptual models of morphological analysis have been created \citep{usmanov2014conceptual}, extensive morpheme databases have been compiled, including 81 prefixes, 76,539 roots, and 128,760 postfixes \citep{dovudov2018computer}, frequency morpheme dictionaries \citep{usmanov2010frequency} and statistical portraits of Tajik text \citep{usmanov2015letter,usmanov2016bigram,kosimov2021syllable} have been developed. Based on these studies, automatic text processing systems TajLINGVO \citep{khudoyberdiev2023modeling}, speech corpora \citep{khudoyberdiev2025modeling}, and open-source toolkits TajikNLP \citep{arabov2026tajiknlp} have been built, as well as the largest corpora of the Tajik language — the Tajik Web Corpus with over 1.1 billion characters \citep{arabov2026tajikweb} and the Tajik National Corpus (NKTJ) with 58.4 million word occurrences \citep{tajikcorpus2026}.

Significant results have also been achieved in adapting neural network architectures for Tajik. POS tagging benchmarking based on multilingual transformers has been conducted \citep{arabov2026benchmarking}, a systematic analysis of subword tokenizer efficiency \citep{arabov2026subword}, a comparative study of parameter-efficient fine-tuning (PEFT) methods for Tajik text generation \citep{arabov2026benchmarkingpeft}, and Tajik-specialized models of the Soro family based on continued pre-training have been created \citep{liashkov2026soro}.

However, despite the availability of the aforementioned resources and methods, the task of creating an electronic explanatory dictionary of Tajik using large language models remains unresolved. Existing works leave undefined the dictionary's architectural model, subword tokenization strategy considering agglutination, fine-tuning methodology under data scarcity conditions, and the evaluation system for generated entries.

This paper represents the first stage of a systematic solution to this problem. The aim of the work is to develop a conceptual framework for an electronic explanatory dictionary of the Tajik language based on large language models. To achieve this goal, the following tasks are addressed:

\begin{enumerate}
\item systematization of existing linguistic, statistical, and infrastructural developments for Tajik relevant to creating a lexicographic resource;
\item analysis of current approaches to adapting LLMs for low-resource languages with emphasis on applicability to Tajik;
\item justification of architectural solutions and methods for creating an electronic explanatory dictionary;
\item development of a conceptual dictionary architecture integrating modules for morphological analysis, lemmatization, semantic clustering, and dictionary entry generation.
\end{enumerate}

The scientific novelty of the work lies in proposing for the first time a holistic conceptual architecture for an explanatory dictionary of Tajik that unifies classical lexicographic description methods, results of statistical analysis of Tajik text, and generative capabilities of large language models. The proposed architecture represents the first stage in creating and subsequently experimentally evaluating a full-featured explanatory dictionary of Tajik.

The practical significance of the study lies in forming a methodological foundation for developing an electronic explanatory dictionary of Tajik, which can serve both as a lexicographic tool and as a foundational resource for a wide range of applied NLP tasks: machine translation, automatic summarization, sentiment analysis, question-answering systems, and others.

The paper is addressed to specialists in computational linguistics, lexicography, developers of natural language processing systems, and researchers working on digital development issues for low-resource languages.


\section{Related Work}

\subsection{Morphological Foundation of the Tajik Language}

Fundamental research on Tajik morphology was conducted within the scientific school formed around the work of Z.D. Usmanov. The monograph "Morphological Analysis of Tajik Word Forms" \citep{usmanov2015morphological} describes inflectional categories and grammatical features of parts of speech, provides classification of affix types and word forms, examines the transformation of parts of speech when affixes are attached, and proposes algorithms for automatic morphological analysis. The conceptual model of the morphological analyzer is presented as a flowchart reflecting the functioning of individual subsystems and the system as a whole \citep{usmanov2014conceptual}.

The most comprehensive morpheme database of Tajik was compiled in the dissertation research of G.M. Dovudov \citep{dovudov2018computer}, containing 81 prefixes, 76,539 roots, and 128,760 postfixes. The work proposes a classification of affix types—inflectional, derivational, and collocational—and a similar classification of word forms. Positional encoding of Tajik word forms and equivalent representation of collocational word forms as sentence fragments were developed. Based on the morphological analyzer, language packages for spell-checking in OpenOffice.org and Microsoft Office were created and are used in organizations of the Republic of Tajikistan.

Frequency characteristics of morphemes were established in \citep{usmanov2010frequency}, where the frequency of morpheme structures and the composition of the most common morphs are determined through processing a large volume of textual information. Research on prefixes of the Tajik literary language \citep{usmanov2009prefix}, using combinatorial and statistical methods, establishes a list of elementary and aggregate prefixes with their repetition frequencies. Statistics of parts of speech in Tajik \citep{dovudov2012statistics}, based on semi-automatic processing of a text corpus, determines the frequency of occurrence of parts of speech of Tajik word roots.

Classification of Tajik words by formation types was continued in the work of Madibragimov and Prutskov. Article \citep{madibragimov2020classification} presents a classification of nouns (5 types, 12 subtypes); \citep{madibragimov2022intermediate} provides classification results for the following parts of speech: nouns — 5 types, 12 subtypes; verbs — 9 types, 2 subtypes; adjectives — 5 types, 2 subtypes; pronouns — 5 types. Separate works are devoted to classification of verbs \citep{madibragimov2023verbs}, adjectives and pronouns \citep{madibragimov2021adjectives}. The authors developed an Internet application for word form generation based on the created linguistic database \citep{madibragimov2023internet}.

The syllable structure of Tajik was studied in \citep{usmanov2006syllable}, where statistical processing of a representative sample from various Tajik texts revealed 2,978 different syllable structures of words and 6 different syllable structures.

\subsection{Statistical Portraits of Text and Frequency Characteristics}

A significant number of works are devoted to establishing statistical patterns of Tajik text. In \citep{usmanov2015letter}, it is established that letter frequency in Tajik in classical and modern poetry and prose is statistically indistinguishable; data on letter frequency and average information per alphabet character are provided. The study of bigram frequency \citep{usmanov2016bigram} shows statistical indistinguishability of distributions in classical and modern poetry, as well as in modern prose; a list of bigrams with the highest frequency of occurrence is provided.

The task of author recognition of a text fragment based on syllable frequency is considered in \citep{kosimov2021syllable}, where each work is associated with a digital portrait — the distribution of syllable frequencies. Using the gamma classifier and the nearest neighbor method, it is possible to identify the author from fragments ranging from 7,000 words to 20 words. The monograph \citep{usmanov2022software} presents the development of an authorship recognition methodology that ensures increased accuracy and reduced time costs.

Studies on the euphony of Tajik words are devoted to works \citep{pirov2023euphony,pirov2022euphony,usmanov2015euphonometry}. In \citep{pirov2023euphony}, based on trigrams and using the gamma classifier, the problem of recognizing word euphony is solved; a similar study based on unigrams is presented in \citep{pirov2022euphony}. A systematic approach to forming the foundations of euphonometry of words is outlined in \citep{usmanov2015euphonometry}.

In \citep{usmanov2015shahname}, two formulas are proposed for describing the correlation of the numbers of word forms and word usages in the work of A. Firdousi ("Shahname"). The diversity of word-form anagrams in English, Russian, Tajik, and Uzbek languages is investigated in \citep{usmanov2022anagrams}; automatic search and statistical patterns of the set of anagrams are presented in the monograph \citep{usmanov2020anagrams}.

The gamma classifier as a tool for statistical analysis of texts is described in detail in the review \citep{usmanov2021gamma}; features of its application for recognizing homogeneous objects are considered in \citep{usmanov2021homogeneous}. The effectiveness of consonant writing in Tajik script is investigated in \citep{usmanov2019consonant}. Issues of automatic recognition of authorship and styles of works of Tajik-Persian fiction are considered in \citep{usmanov2020authorship}. Testing of the gamma classifier tuned for language recognition of works based on the Latin alphabet is described in \citep{usmanov2021gamma_testing}.

\subsection{Automatic Text Processing Systems and Corpus Infrastructure}

In \citep{khudoyberdiev2023modeling}, a methodology for forming text information processing processes in the TajLINGVO system is developed, the logical structure of the system is proposed with a detailed description of subprocesses, as well as a functional model based on UML. The effectiveness of the models is confirmed in the development of a computer thesaurus, spell-checking, speech synthesis, and machine translation for Tajik.

A mathematical model of the speech corpus processing process for Tajik is presented in \citep{khudoyberdiev2025modeling}; methods of corpus formation, interaction of subjects and the system, and stages of database design for collecting and processing audio recordings are described. Issues of machine translation and its relationship with literary translation are considered in \citep{khudoyberdiev2021translation}; a classification of machine translators by functionality is proposed. An overview of neural network technologies as applied to Tajik linguistics is presented in \citep{karimova2023neural}. Algorithms for syntactic analysis of simple sentences in Tajik are developed in \citep{karimova2025syntax}.

The most significant corpus resources are: the Tajik National Corpus (NKTJ) with 58.4 million word occurrences and 96\% automatic parsing coverage with grammatical information for each word form \citep{tajikcorpus2026}; the Tajik Web Corpus, containing 319,298 documents (168.5 million words, ~1.11 billion characters), which is the largest open corpus of Tajik \citep{arabov2026tajikweb}; the Tajik--Persian parallel corpus of 328,253 aligned sentences \citep{arabov2026tajpersparallel}.

In \citep{arabov2026subword}, a comparative analysis of five subword tokenization models (BPE, WordPiece, Unigram) on a corpus of over 33 million tokens is conducted; strengths and weaknesses of different approaches are identified and the most effective tokenization strategies for Tajik are determined. In \citep{arabov2026tajiknlp}, the open-source library TajikNLP is presented, implementing a complete Tajik text processing pipeline: cleaning, normalization, tokenization, morpheme segmentation, POS tagging, stemming, lemmatization. The library includes a unified morphology mechanism with controlled and deep analysis modes, a lexicon-based sentiment analyzer, and pre-trained Word2Vec/FastText embeddings.

The creation of a multiformat text corpus for training modern language models is described in \citep{arabov2025multiformat}; results on collecting and processing textual data in Tajik from various sources are presented. For automatic spelling checking, the TajSpell system has been developed \citep{soliev2021tajspell}. Other corpus resources include the unified morphological corpus \citep{tajikcorpus2026unified}, the POS-tagged corpus \citep{tajikpos2025}, the full Tajik--Persian parallel corpus \citep{tajpersparallel2026full}, and the dataset of Tajik names \citep{tajiknames2026}.

\subsection{Large Language Models for the Tajik Language}

In \citep{arabov2025developing}, the structural reasons for the absence of Tajik in modern LLM ecosystems are analyzed. Three interrelated domains limiting technology development are identified: data availability and quality, linguistic representation, and research infrastructure. Special attention is paid to the divergence between the classical linguistic proximity of Tajik and Persian and functional technological compatibility. A conceptual framework and research program for corpus construction, linguistic preprocessing, and safe model adaptation are formulated. Issues of data sovereignty in the era of LLMs for the Republic of Tajikistan are considered in \citep{arabov2026sovereignty}.

In \citep{arabov2026benchmarkingpeft}, based on the Tajik Web Corpus \citep{arabov2026tajikweb}, a benchmarking of 17 model configurations of different classes using three fine-tuning strategies is conducted. The best result was achieved by the Mistral 7B model with QLoRA (rank 16), reaching an average perplexity of 5.03. It is established that full fine-tuning for small GPT-2 models leads to catastrophic forgetting, while parameter-efficient methods allow preservation of generative capability in Tajik.

In \citep{arabov2026benchmarking}, the first benchmark for POS tagging of Tajik on the TajPersParallel corpus (about 44,000 dictionary entries) is presented. The best result (weighted F1 = 0.62) was achieved by the mBERT model with LoRA fine-tuning. It is established that in the absence of syntactic context, all models experience difficulties in resolving morphological ambiguity; zero-shot evaluation showed the greatest typological similarity of Tajik with Persian and Russian.

The study of Tajik-Persian transliteration \citep{arabov2026transliteration} presents results of a comparative analysis of six model classes. The ByT5 model achieved chrF++ of 87.4 ± 0.1 for Tajik $\to$ Farsi and 80.1 ± 0.2 for the reverse direction. It is established that for accurate transliteration, architectures operating at the byte or character level are more effective than multilingual Seq2Seq models based on subword tokenization. Similar results based on a parallel lexical corpus of 52,152 pairs were obtained in \citep{kurbonovich2026character}.

In \citep{arabov2026tajperslexon}, TajPersLexon is presented — a parallel lexical resource of 40,112 word and short phrase pairs; neural and search models achieve 98--99\% top-1 accuracy.

In \citep{liashkov2026soro}, the Soro family is presented — Tajik-specialized LLMs based on Gemma 3 with continued pre-training on a 1.9 billion token corpus and instruction fine-tuning on 40,000 examples. A set of benchmarks for Tajik is proposed, covering general knowledge, linguistic competence, and examination tasks. Soro significantly outperforms base Gemma 3 models of the same size.

In \citep{arnob2026one}, an evaluation of open multilingual LLMs on three regional variants — Persian, Dari, and Tajik — is conducted. Serious performance discrepancies are identified: almost all models show catastrophic degradation on Tajik script (drop to 1.0 BLEU-100 score). The obtained results highlight a critical "script barrier" in modern open multilingual LLMs.

A practical guide to the modern NLP pipeline with emphasis on low-resource languages is presented in \citep{arabov2026nlp}; the material includes linguistic resources for Tajik and Tatar languages — subword tokenizers, embeddings, lexical databases, and transliteration benchmarks. A comparative analysis of methods for modeling semantic representations of words for Tajik under limited language resources is conducted in \citep{arabov2025semantic}.

\subsection{LLM Lexicography for Low-Resource Languages: General Approaches}

Global experience in recent years demonstrates a consistent trend toward the application of large language models (LLMs) in lexicographic practice, including for languages with limited digital resources.

One of the first demonstrations of the possibility of automatic generation of dictionary definitions was the work of Bear and Cook \citep{bear2021cross}, who used BPE subword representations to build English definitions for the endangered polysynthetic language Wolastoqey, outperforming baseline methods on the BLEU metric.

With the spread of ChatGPT, a qualitatively new stage began. Jakub{\'\i}{\v{c}}ek and Rundell \citep{jakubicek2023chatgpt}, using the example of a fully automatically created English dictionary of 99 entries, showed that ChatGPT demonstrates a level close to state-of-the-art for a number of dictionary entry components, but has significant limitations. De Schryver \citep{schryver2023generative}, summarizing ten works on using ChatGPT in lexicography, confirmed that a carefully crafted prompt allows generating dictionary entries comparable to the best examples of traditional lexicography.

However, when transitioning to low-resource languages, commercial LLMs demonstrate significantly more modest results. Ojo and Ogueji \citep{ojo2023commercial} showed that on eight African languages, the quality of machine translation and text classification remains low, which stimulated the creation of specialized solutions.

To fill lexicographic gaps, a number of specialized solutions have been proposed. The GUIDE tool \citep{janetzki2024guide} — a language-independent system based on graph neural networks — creates semantic domain dictionaries using parallel corpora of Bible translations. In zero-shot mode, GUIDE predicts an average of 2,400 dictionary entries with approximately 60\% accuracy.

For Southeast Asian languages, the SeaLLMs family \citep{nguyen2024seallms} was created with continued pre-training and instruction tuning. The models outperformed ChatGPT-3.5 in non-Latin scripts (Thai, Khmer, Lao, Burmese).

For the African continent, the Cheetah model \citep{adebara2024cheetah} was developed, supporting 517 languages and outperforming counterparts in five of six text generation tasks. The AfriqueLLM family \citep{yu2026afriquellm} was also created — open LLMs adapted to 20 African languages through continued pre-training on 26 billion tokens. It was shown that data composition is the main success factor.

For Indian languages, the LexGen model \citep{maheshwari2025lexgen} was proposed, generating domain-oriented dictionaries in multi-domain mode. A reference set of over 75,000 translation pairs in eight domains was created; zero-shot and few-shot experiments confirmed the model's ability to generalize to unseen domains and languages.

Practical experience in extracting structured lexical information using LLMs is presented in \citep{jumashev2025structured}, where GPT-4o was used to convert entries from the Russian-Kyrgyz dictionary of Yudakhin into JSON schema. The combination of few-shot learning and fine-tuning achieved 92.7\% accuracy.

The task of generating usage examples for bilingual dictionaries was investigated by Merx et al. \citep{merx2024generating} on languages of varying resource levels (French, Indonesian, Tetun). It was revealed that example quality noticeably degrades for low-resource Tetun, and perplexity can serve as a proxy metric for typicality.

Experience in dialect lexicography with RAG is described by St{\"o}ckle et al. \citep{stockle2025llm} on the material of historical Bavarian dialects of Austria. A modular pipeline for automated dictionary creation with human intervention capability is presented by Widmann \citep{widmann2025pipeline}.

Rab{\'e} et al. \citep{rabe2025taboo} developed a semi-automatic methodology for compiling a list of taboo constructions for Afrikaans, based on frequency analysis and expert verification. Lugli \citep{lugli2025mangalam} described the first corpus-oriented dictionary of Buddhist Sanskrit, where generative LLMs were used for semantic annotation.

Multilingual definition modeling was investigated by Marrese-Taylor et al. \citep{marrese2025multilingual}, who showed that multilingual LLMs are capable of generating definitions at a level comparable to English. Lu et al. \citep{lu2026dictionary} proposed the Dictionary Insertion Prompting (DIP) method, improving LLM reasoning by inserting English equivalents of words into non-English queries.

Although the work of Kassab et al. \citep{kassab2025automated} is devoted to creating a dataset for NER in Russian, the semi-automatic pipeline with ensemble voting proposed therein, which improved annotation accuracy by 28\%, is of interest as a methodological example for forming high-quality training samples in dictionary construction.

To systematize the considered approaches, Table \ref{tab:comparison} provides a comparison of key projects.

\begin{table}[htbp]
\centering
\caption{Comparison of LLM Lexicography Approaches for Low-Resource Languages}
\label{tab:comparison}
\begin{tabular}{p{3cm}p{3cm}p{4cm}p{4cm}}
\toprule
\textbf{Project/Model} & \textbf{Target Languages} & \textbf{Main Method} & \textbf{Key Result} \\
\midrule
Bear \& Cook (2021) \citep{bear2021cross} & Wolastoqey & BPE + Seq2Seq & Advantage over baseline methods on BLEU \\
GUIDE (2024) \citep{janetzki2024guide} & 20+ languages & Graph Neural Networks & 60\% accuracy, ~2,400 entries \\
SeaLLMs (2024) \citep{nguyen2024seallms} & Thai, Khmer, Lao, Burmese & Continued pre-training & Outperforms ChatGPT-3.5 \\
Cheetah (2024) \citep{adebara2024cheetah} & 517 African languages & Continued pre-training & Outperforms counterparts in 5/6 tasks \\
LexGen (2025) \citep{maheshwari2025lexgen} & 6 Indian languages, 8 domains & Domain layers + routing & Zero-shot and few-shot; >75K pairs \\
Jumashev et al. (2025) \citep{jumashev2025structured} & Kyrgyz & GPT-4o + fine-tuning & 92.70\% accuracy, 93.56\% F1 \\
Merx et al. (2024) \citep{merx2024generating} & French, Indonesian, Tetun & Example generation & Quality drops for low-resource \\
St\"ockle et al. (2025) \citep{stockle2025llm} & Bavarian dialects & RAG + semantic classification & Quality confirmed by experts \\
Widmann (2025) \citep{widmann2025pipeline} & Language-independent & Modular pipeline & Reproducibility and quality control \\
AfriqueLLM (2026) \citep{yu2026afriquellm} & 20 African languages & CPT on 26B tokens & Data composition is main factor \\
\textbf{This work} & \textbf{Tajik (Cyrillic)} & \textbf{Morphological base + embeddings + PEFT LLM} & \textbf{Proposed architecture (pending experimental evaluation)} \\
\bottomrule
\end{tabular}
\end{table}

As can be seen from the table, the considered projects cover African, Southeast Asian, Indian languages, and European dialects, but none address Tajik. The proposed architecture is distinguished by the integration of a formal morphological model (76,539 roots, 128,760 postfixes) and extensive corpus infrastructure (Tajik Web Corpus ~168.5 million words, NKTJ ~58.4 million word occurrences).

Thus, the global arsenal of approaches to LLM lexicography for low-resource languages covers all stages — from generating individual definitions to fully automated pipelines with expert control. At the same time, none of the mentioned initiatives address Tajik. This circumstance highlights the existing gap and simultaneously allows reliance on proven architectural solutions. Unlike many African or Southeast Asian languages, Tajik has accumulated extensive corpus data and a well-verified morphological model, making it an ideal candidate for adapting these methods.

\subsection{Justification of the Novelty of This Study}

The conducted review shows that to date, a solid scientific base has been formed for creating an electronic explanatory dictionary of Tajik: fundamental foundations of morphological analysis have been developed \citep{usmanov2015morphological,usmanov2014conceptual,dovudov2018computer}, frequency dictionaries and statistical portraits of text have been created \citep{usmanov2010frequency,usmanov2015letter,usmanov2016bigram,kosimov2021syllable}, automatic text processing systems \citep{khudoyberdiev2023modeling} and open toolkits \citep{arabov2026tajiknlp} have been built, large corpora \citep{arabov2026tajikweb,tajikcorpus2026} have been created, experiments on adapting LLMs for Tajik using parameter-efficient fine-tuning \citep{arabov2026benchmarkingpeft} have been conducted, and Tajik-specialized models \citep{liashkov2026soro} have been created.

However, the task of creating a comprehensive electronic explanatory dictionary of Tajik using large language models remains unresolved. Existing works lack a holistic conceptual architecture defining:

\begin{itemize}
\item the structure of the dictionary and the interaction of its modules;
\item tokenization methods considering the agglutinative structure of Tajik;
\item the LLM fine-tuning strategy under conditions of limited annotated data;
\item data sources for forming the training corpus of the dictionary;
\item criteria for evaluating the quality of generated dictionary entries.
\end{itemize}

This work aims to fill the identified gap and represents the first stage of a systematic solution to the task of creating and subsequently experimentally evaluating an electronic explanatory dictionary of Tajik based on LLMs.


\section{Proposed Architecture of the Electronic Explanatory Dictionary}

\begin{figure}[htbp]
\centering
\includegraphics[width=0.85\textwidth]{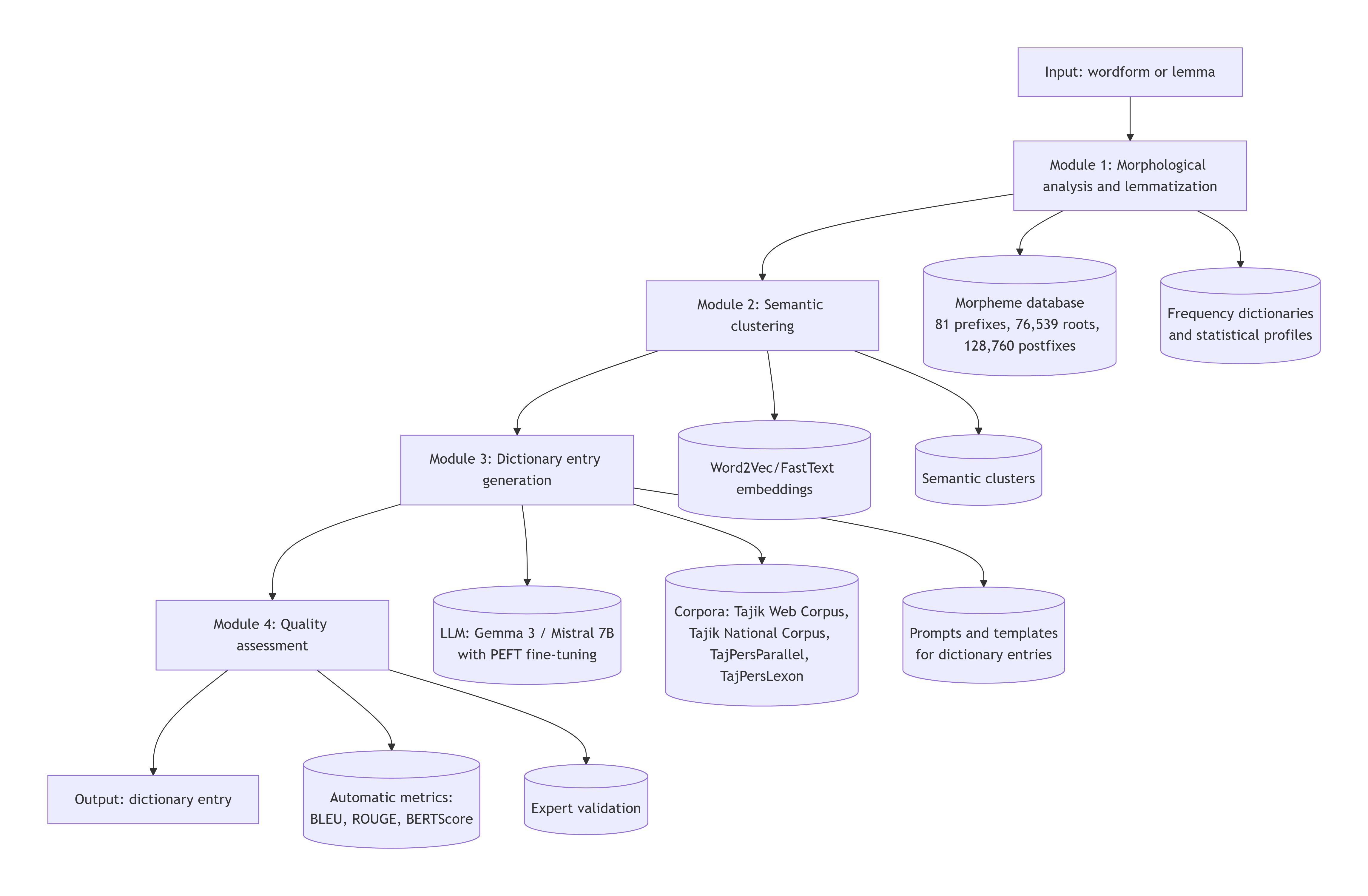}
\caption{Generalized architecture of the LLM-based electronic explanatory dictionary of Tajik}
\label{fig:architecture}
\end{figure}

Based on the systematization of existing linguistic, statistical, and infrastructural developments for the Tajik language (Sections 2.1--2.3), as well as taking into account the global experience in applying large language models in lexicographic practice for low-resource languages (Section 2.5), this section proposes a conceptual architecture for an electronic explanatory dictionary of the Tajik language. The architecture is built as a multi-stage processing pipeline integrating four main functional modules: morphological analysis and lemmatization, semantic clustering, LLM-based dictionary entry generation, and quality assessment. Each module relies on existing corpus resources and tools, as well as methodological solutions proven in global practice. The proposed architecture is not rigidly deterministic: it allows variation of components depending on available computational resources and the target tasks of the dictionary. The generalized data flow diagram of the proposed architecture is shown in Figure \ref{fig:architecture}, which sequentially reflects all processing stages — from word form input to the output of a ready dictionary entry with a feedback loop for iterative quality improvement.

\subsection{Morphological Analysis and Lemmatization Module}

The morphological analysis module performs two key functions: reducing the input word form to its base form (lemma) and extracting grammatical information necessary for subsequent dictionary entry generation. As a foundation for module implementation, fundamental works on Tajik morphology are used \citep{usmanov2014conceptual,dovudov2018computer,usmanov2010frequency}, which established an extensive morpheme database containing 81 prefixes, 76,539 roots, and 128,760 postfixes. The classification of affix types proposed in these works — inflectional, derivational, and collocational — allows sequential identification of morpheme boundaries and determination of grammatical characteristics of the word form. The module's algorithm includes tokenization and normalization of the input word form with conversion to standard Cyrillic representation, positional encoding of the word form using the method developed in \citep{dovudov2018computer}, which allows unambiguous identification of morpheme positions, matching against the morpheme database with sequential extraction of prefixes, roots, and postfixes based on pre-compiled dictionaries \citep{usmanov2009prefix}, determination of part of speech and grammatical categories based on statistical data on the frequency of parts of speech of Tajik word roots \citep{dovudov2012statistics}, and final lemmatization — conversion of the word form to its base dictionary form. For word forms not present in the morpheme database, a morphological inference mechanism is provided based on the classification of words by formation types developed in \citep{madibragimov2020classification,madibragimov2022intermediate,madibragimov2023verbs,madibragimov2021adjectives}. This classification covers nouns (5 types, 12 subtypes), verbs (9 types, 2 subtypes), adjectives (5 types, 2 subtypes), and pronouns (5 types), allowing morphological analysis to be extended to a significant portion of unknown word forms. The syllable structure of the word \citep{usmanov2006syllable} is taken into account at the phonetic normalization stage. The module's output is a structured record containing the lemma, part of speech, grammatical characteristics (gender, number, case, tense, person, etc.), and morpheme composition. This structure is passed to the semantic clustering module and the dictionary entry generation module.

\subsection{Semantic Clustering Module}

The semantic clustering module is designed to group lexemes by semantic fields and thematic classes. The need for such a module stems from the fact that an explanatory dictionary should not only provide definitions of individual words but also reflect systematic semantic relationships between them: synonymy, antonymy, hypo-hyperonymic relations, thematic proximity. As a foundation for semantic clustering, pre-trained Word2Vec and FastText embeddings included in the TajikNLP open-source library are used \citep{arabov2026tajiknlp}. The choice of embeddings as the primary word representation is due to their ability to reflect semantic proximity based on the distributional hypothesis, which has been confirmed for Tajik in \citep{arabov2026tajiknlp}, where embeddings were trained on a corpus of over 33 million tokens. The clustering algorithm includes vectorization of lemmas — converting each lemma into a vector representation using pre-trained Word2Vec or FastText embeddings, computation of semantic similarity between vectors based on cosine or Euclidean distance, grouping of lexemes into clusters using clustering algorithms (e.g., K-means or hierarchical clustering) followed by expert verification, and final formation of semantic fields by merging clusters into larger thematic groups based on linguistic criteria. The obtained semantic clusters and fields are used in the dictionary entry generation module to include synonyms and antonyms in the entry, indicate the thematic affiliation of the word, and provide usage examples from contexts semantically close to the given lexeme. The integration of semantic clustering into the dictionary architecture allows overcoming the limitation of traditional explanatory dictionaries, which often present words in isolation without explicit indication of systematic semantic relationships.

\subsection{LLM-Based Dictionary Entry Generation Module}

The dictionary entry generation module is the central component of the proposed architecture. Its task is to generate a comprehensive dictionary entry based on the lemma, its grammatical characteristics, and semantic cluster, including definition, usage examples, grammatical labels, synonyms, and antonyms. As the base architecture, open-weight generative large language models are proposed, which have shown effectiveness for Tajik in recent benchmarks. The results of \citep{arabov2026benchmarkingpeft} demonstrate that Mistral 7B family models with parameter-efficient fine-tuning (QLoRA, rank 16) achieve an average perplexity of 5.03 on Tajik text generation, which is the best result among 17 tested configurations. At the same time, full fine-tuning of small GPT-2 models leads to catastrophic forgetting, while parameter-efficient methods allow preservation of generative capability in Tajik. The Soro model created in \citep{liashkov2026soro}, based on the Gemma 3 architecture with continued pre-training on a 1.9 billion token corpus and instruction fine-tuning on 40,000 Tajik examples, demonstrates significant superiority over the base model on Tajik tests covering general knowledge, linguistic competence, and examination tasks. Considering these results, the main candidates for the base model are Gemma 3 in versions 2B or 7B as a model with continued pre-training on the Tajik corpus \citep{liashkov2026soro}, and Mistral 7B as a model that showed the best results in the PEFT benchmark for Tajik \citep{arabov2026benchmarkingpeft}. The final choice between Gemma 3 and Mistral 7B will be determined experimentally based on comparison of generation quality on a validation set of 100--200 dictionary entries, with priority given to the model with the best balance between perplexity and expert evaluation.

Given the limited availability of annotated data for Tajik, an approach using Parameter-Efficient Fine-Tuning (PEFT) with LoRA (Low-Rank Adaptation) or QLoRA (Quantized LoRA) methods has been developed, as confirmed by the results of \citep{arabov2026benchmarkingpeft}. The advantages of this approach include minimal computational costs when fine-tuning a small number of parameters instead of the full model, prevention of catastrophic forgetting, and the ability to quickly adapt to different types of dictionary entries. For the training set, it is planned to use existing explanatory dictionaries of Tajik \citep{shukurov2008tajik} in electronic form, parallel corpora and lexical resources \citep{arabov2026tajpersparallel,arabov2026tajperslexon}, as well as synthetically generated ``word -- dictionary entry'' pairs based on existing lexicographic data and methods tested in \citep{jumashev2025structured}.

For dictionary entry generation, a prompt system has been developed, including a base prompt with general instructions for the model on the format and content of the dictionary entry, a structured prompt specifying required fields (definition, grammatical labels, examples, synonyms, antonyms), and a contextual prompt including the semantic cluster to ensure thematic coherence. Approaches to prompting for dictionary entry generation tested in \citep{jakubicek2023chatgpt,schryver2023generative}, including few-shot learning on examples, have been adapted for Tajik. The Dictionary Insertion Prompting (DIP) method \citep{lu2026dictionary}, which improves LLM reasoning by inserting equivalents, can be used when working with Persian parallels. The proposed dictionary entry format includes lemma, part of speech, grammatical labels (gender, number, case, tense, person), definition (meaning explanation), usage examples (2--3 sentences from the corpus or generated), synonyms, antonyms, thematic group, and, optionally, etymology. This structure corresponds to the best examples of traditional lexicography \citep{shukurov2008tajik} and is adapted for digital representation.

To demonstrate the interaction of modules, we provide an end-to-end example of processing the word form \tj{китобҳоямро} (``my books'' as a direct object). The choice of this word form is due to its agglutinative complexity: it contains a root, a plural suffix, a 1st person singular possessive affix, and a case marker for the accusative case, which allows testing all main stages of morphological analysis. The input to the morphological analysis module is the word form \tj{китобҳоямро}. Based on the morpheme database \citep{dovudov2018computer}, morpheme segmentation is performed: \tj{китоб} (root, noun) + \tj{-ҳо} (plural marker) + \tj{-ям} (1st person singular possessive affix) + \tj{-ро} (accusative case marker). Using the classification of formation types \citep{madibragimov2020classification}, the word form is identified as a noun of type 1, subtype 1.1 (consonant stem). As a result, grammatical features are formed: lemma = \tj{китоб}, part of speech = noun, number = plural, possessiveness = 1st person singular, case = accusative. Next, the semantic clustering module converts the lemma into a vector representation using pre-trained FastText embeddings (dimension 300, trained on a corpus of 33 million tokens \citep{arabov2026tajiknlp}) and computes the cosine distance to cluster centroids. The lemma is assigned to the semantic cluster ``book, reading, library, publication,'' and synonyms from the same cluster are identified: \tj{асар, нашрия, дастнавис}. At the final stage, a prompt containing the lemma, grammatical labels, and semantic cluster information is fed to the LLM (e.g., Mistral 7B with QLoRA, rank 16, fine-tuned on the Tajik Web Corpus according to the protocol of \citep{arabov2026benchmarkingpeft}). As a result, the following dictionary entry is generated:

\begin{quote}
\tj{Китоб} (noun, pl. \tj{китобҳо}) — a collection of printed or manuscript sheets containing scientific, literary, or informational texts. Usage examples: \tj{Вай ҳар рӯз се соат китоб мехонад}; \tj{Китобҳоямро ба китобхонаи мактаб супурдам}. Synonyms: \tj{асар, нашрия, дастнавис}. Thematic group: \tj{маориф, фарҳанг, китобдорӣ}.
\end{quote}

This example illustrates how a lexicographically complete entry is synthesized from morphologically annotated input, semantic information based on distributional embeddings, and the generative capabilities of LLMs trained on Tajik corpora.

\begin{figure}[htbp]
\centering
\includegraphics[width=\textwidth]{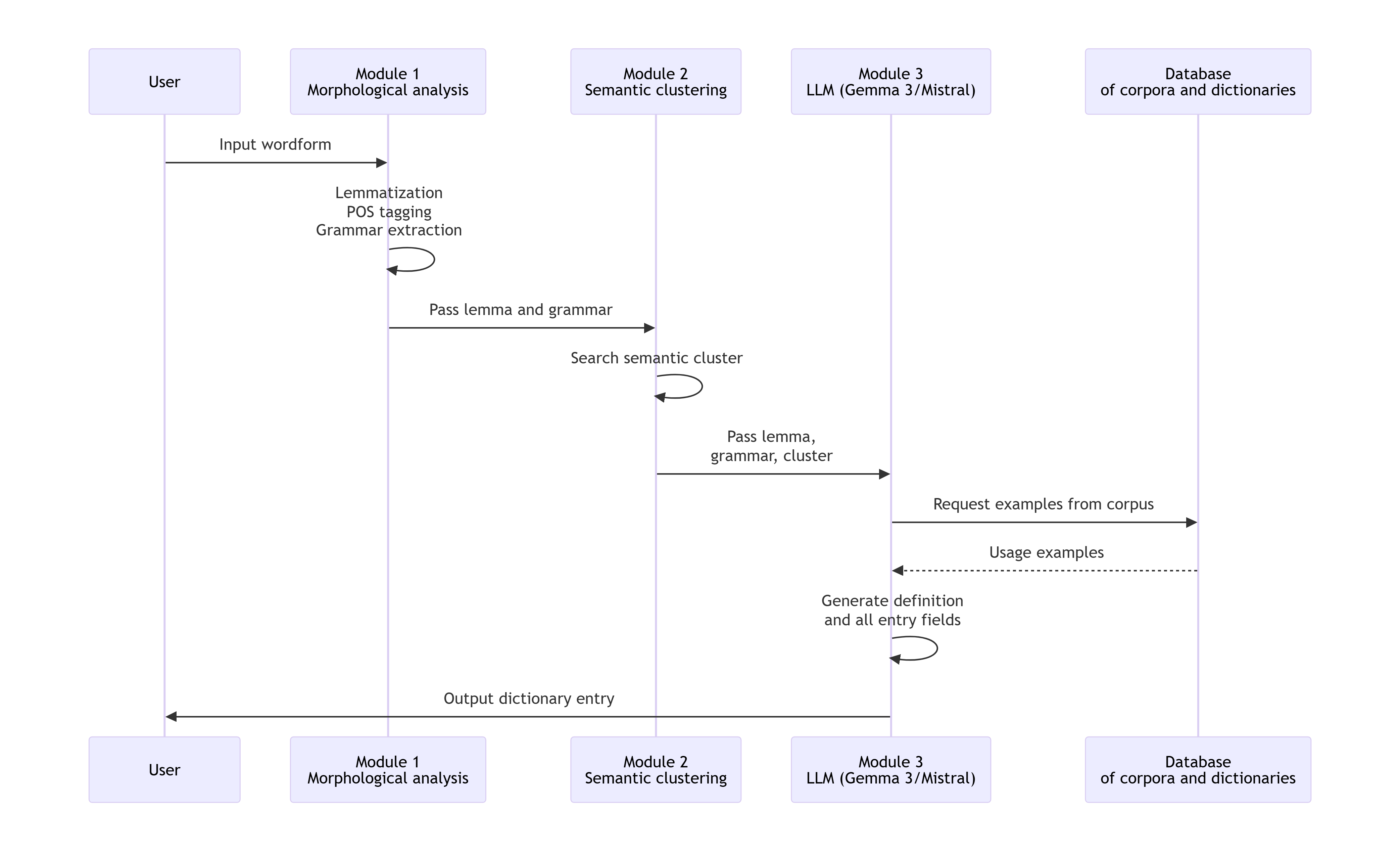}
\caption{Sequence diagram of operations for dictionary entry generation}
\label{fig:sequence}
\end{figure}

The sequence diagram of operations for dictionary entry generation is shown in Figure \ref{fig:sequence}, which reflects the full cycle of interaction between the user, morphological analysis and semantic clustering modules, LLM, and the corpus and dictionary database.

\subsection{Quality Assessment Module}

The quality assessment module is designed to verify generated dictionary entries and provides both automatic control and expert validation capability. The need for this module stems from the fact that generative LLMs, despite high quality, can make factual errors, hallucinate, or produce incorrect formulations \citep{jakubicek2023chatgpt}. For automatic quality assessment of generated entries, the following metrics are proposed: BLEU (Bilingual Evaluation Understudy) for n-gram similarity evaluation with reference dictionary entries, ROUGE (Recall-Oriented Understudy for Gisting Evaluation) for evaluating recall and precision on n-grams, METEOR (Metric for Evaluation of Translation with Explicit ORdering) as a metric that accounts for synonymy and word order, and BERTScore as a metric based on BERT embeddings for semantic similarity evaluation. BERTScore is proposed as the primary metric since it better correlates with expert evaluation for semantically rich tasks such as lexicography, unlike BLEU, which is sensitive to lexical overlap and does not account for semantic synonyms and paraphrases. As reference entries, existing explanatory dictionaries of Tajik \citep{shukurov2008tajik} are used, as well as, where available, entries from Persian lexicographic sources adapted to the Tajik context.

\begin{figure}[htbp]
\centering
\includegraphics[width=\textwidth]{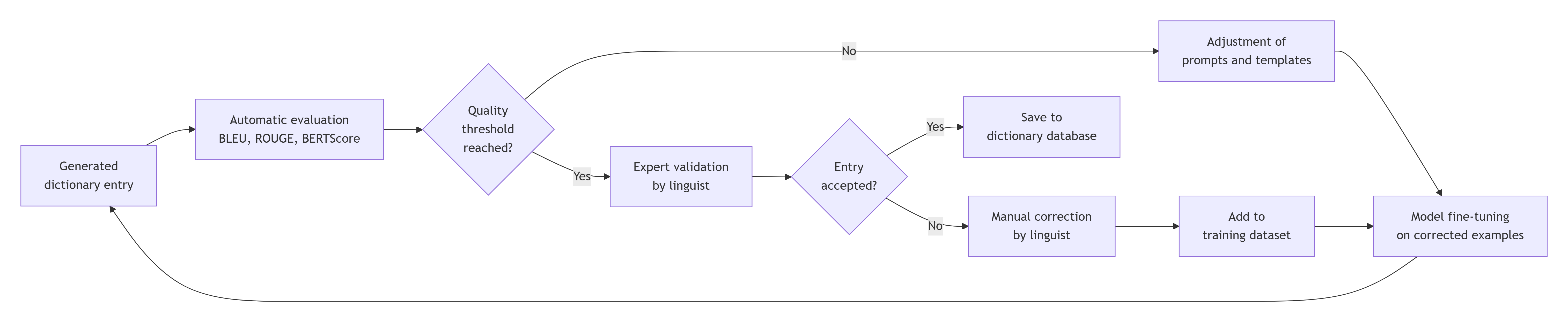}
\caption{Cycle of evaluation and quality improvement of dictionary entries}
\label{fig:quality}
\end{figure}

To ensure high quality of the final product, expert validation of generated entries by qualified linguists is provided. The validation procedure includes verification of definition correctness, assessment of example naturalness, verification of grammatical labels, and correction of semantic relations (synonyms, antonyms). The modular pipeline with human intervention capability at each stage, tested in \citep{widmann2025pipeline}, has been adapted for Tajik. The evaluation results (both automatic and expert) are used for iterative improvement: adjustment of prompts and generation templates, model fine-tuning on corrected examples, and refinement of semantic clusters. The cycle of evaluation and quality improvement of dictionary entries is shown in Figure \ref{fig:quality}, which reflects all stages from entry generation to its storage in the dictionary database after passing automatic and expert control.

\subsection{Data Sources and Infrastructure Implementation}

The proposed architecture relies on the following corpus resources and tools created for the Tajik language. The Tajik Web Corpus, containing 319,298 documents, 168.5 million words, and over 1.1 billion characters \citep{arabov2026tajikweb}, serves as the foundation for context extraction, LLM fine-tuning, and formation of usage examples. The Tajik National Corpus (NKTJ) with 58.4 million word occurrences and 96\% automatic parsing coverage \citep{tajikcorpus2026} provides morphological annotation and verification of grammatical characteristics. The Tajik--Persian parallel corpus of 328,253 aligned sentences \citep{arabov2026tajpersparallel} and the lexical resource TajPersLexon of 40,112 word and short phrase pairs \citep{arabov2026tajperslexon} provide cross-linguistic parallels for synonymy, translation, and cross-script alignment. The morpheme database containing 81 prefixes, 76,539 roots, and 128,760 postfixes \citep{dovudov2018computer} ensures morphological analysis and lemmatization. The open-source Python library TajikNLP \citep{arabov2026tajiknlp} implements a complete text processing pipeline, including tokenization, POS tagging, and lemmatization. Infrastructure-wise, the system can be implemented as a web application with a Python backend using Hugging Face Transformers, PEFT, and SentencePiece libraries, and a user interface providing word form input and structured dictionary entry output. The architecture allows both local deployment on GPU servers and the use of cloud solutions.

\subsection{Discussion of Limitations and Implementation Prospects}

The proposed architecture has a number of limitations that must be taken into account at the implementation stage. Generation quality depends on training data: despite the availability of large corpora, labeled ``word -- dictionary entry'' pairs for Tajik are not available in sufficient volume, which requires the development of methods for synthetic formation of training sets (as described in Section 3.3) or the use of cross-linguistic transfer. To minimize the impact of errors in corpora, ensemble filtering is proposed: the entry is generated based on several different context samples, and the final version is selected based on consensus on model confidence metrics. The morphological complexity of Tajik, its agglutinative structure, and high variability of word forms create additional challenges for subword tokenization. However, the results of \citep{arabov2026subword} show that BPE and Unigram with properly selected parameters can effectively handle this task. As shown in \citep{arnob2026one}, multilingual LLMs demonstrate catastrophic degradation on Tajik Cyrillic script, which confirms the need for Tajik-specialized models  or fine-tuning of base models on the Tajik corpus. LLM fine-tuning requires significant computational power, but the use of parameter-efficient methods \citep{arabov2026benchmarkingpeft} significantly reduces these requirements. Fully automatic generation without expert control does not guarantee lexicographic quality, which requires the development of effective semi-automatic pipelines \citep{widmann2025pipeline}. Despite these limitations, the proposed architecture relies on solutions tested in global practice, adapted to the specific features of Tajik and existing developments in its computational linguistics. The laid conceptual foundation determines the next steps for practical prototype implementation, experimental comparison of alternative models, and user interface development, which will constitute the content of subsequent sections of this work.


\section{Conclusion}

In this work, a conceptual architecture for an electronic explanatory dictionary of the Tajik language based on large language models has been proposed. The relevance of the study is due to the absence in Tajik of a comprehensive digital lexicographic resource comparable in functionality to explanatory dictionaries of high-resource languages, as well as the insufficient adaptation of modern natural language processing technologies to low-resource language systems \citep{arabov2025developing}. The conducted systematic review has shown that, despite the presence of fundamental developments in computational linguistics for Tajik, including morphological analysis \citep{usmanov2014conceptual,dovudov2018computer,usmanov2010frequency}, frequency dictionaries and statistical portraits \citep{usmanov2016bigram,usmanov2015letter,kosimov2021syllable}, and corpus resources \citep{arabov2026tajikweb,tajikcorpus2026}, the task of creating a comprehensive explanatory dictionary using large language models has remained unresolved.

The review of global experience in LLM lexicography for low-resource languages (Section 2.5) allowed identifying proven methodological solutions applicable to Tajik: from generation of individual definitions using subword representations \citep{bear2021cross} to fully automated pipelines with expert control \citep{widmann2025pipeline}. It has been established that existing initiatives cover a wide range of languages — from African \citep{ojo2023commercial,adebara2024cheetah,yu2026afriquellm} to Southeast Asian \citep{nguyen2024seallms} and Indian \citep{maheshwari2025lexgen} — however, none of them addresses Tajik, which highlights the existing gap and simultaneously creates a methodological foundation for adapting proven solutions.

The work proposes an architecture integrating four modules: morphological analysis and lemmatization, semantic clustering, LLM-based dictionary entry generation, and quality assessment. The morphological analysis module relies on a database of 81 prefixes, 76,539 roots, and 128,760 postfixes \citep{usmanov2010frequency} and classification of words by formation types \citep{madibragimov2020classification,madibragimov2022intermediate,madibragimov2023verbs,madibragimov2021adjectives}, which ensures lemmatization and extraction of grammatical information. The semantic clustering module uses pre-trained Word2Vec and FastText embeddings from the TajikNLP library \citep{arabov2026tajiknlp} for grouping lexemes by semantic fields, which allows including synonyms, antonyms, and thematic labels in dictionary entries. The central component is the LLM-based generation module, which proposes the use of Tajik-specialized models of the Soro family \citep{liashkov2026soro} or Mistral 7B with parameter-efficient fine-tuning \citep{arabov2026benchmarkingpeft} — the choice between them will be determined experimentally on a validation set of 100--200 dictionary entries. A multi-level prompt system has been developed, taking into account the lemma, grammatical characteristics, and semantic cluster. The quality assessment module provides automatic control using BLEU, ROUGE, METEOR, and BERTScore metrics, with BERTScore selected as the primary metric due to its better correlation with expert evaluation for semantically rich tasks, as well as expert validation by qualified linguists following a procedure similar to \citep{widmann2025pipeline}. To minimize errors in corpus data, an ensemble filtering mechanism has been proposed, in which the entry is generated based on several context samples with subsequent consensus selection.

The scientific novelty of the work lies in the fact that for the first time a holistic conceptual architecture for an explanatory dictionary of Tajik has been proposed, uniting classical lexicographic description methods, results of statistical analysis of Tajik text, and generative capabilities of large language models. Unlike many African and Southeast Asian low-resource languages, Tajik has accumulated not only an extensive corpus base (including the Tajik Web Corpus and NKTJ) but also a well-verified morphological model, which makes it an ideal candidate for transferring and adapting LLM lexicography methods proven in global practice.

The practical significance of the study lies in forming a methodological foundation for developing a full-featured electronic explanatory dictionary of Tajik. Such a dictionary can serve both as a lexicographic tool and as a foundational resource for a wide range of applied natural language processing tasks: machine translation, automatic summarization, sentiment analysis, question-answering systems, and others.

Prospects for further research include: practical implementation of a prototype dictionary using the selected LLM; experimental comparison of alternative models (Gemma 3 vs. Mistral 7B) on a validation set; development of a user interface for word form input and structured dictionary entry output; formation of a training set of ``word -- dictionary entry'' pairs based on existing dictionaries \citep{shukurov2008tajik} and synthetic methods \citep{jumashev2025structured}; conducting automatic and expert quality assessment of generated entries; and iterative improvement of the system based on evaluation results.

\bibliographystyle{unsrtnat}
\bibliography{references}

\end{document}